\documentclass[sn-mathphys,Numbered,iicol]{sn-jnl}

\usepackage{graphicx}%
\usepackage{multirow}%
\usepackage{array}%
\usepackage{amsmath,amssymb,amsfonts}%
\usepackage{amsthm}%
\usepackage{mathrsfs}%
\usepackage{subcaption}
\usepackage[title]{appendix}%
\usepackage{xcolor}%
\usepackage{textcomp}%
\usepackage{manyfoot}%
\usepackage{booktabs}%
\usepackage{algorithm}%
\usepackage{adjustbox}
\usepackage{algorithmicx}%
\usepackage{algpseudocode}%
\usepackage{listings}%
\usepackage{caption}%
\usepackage{dirtytalk}%
\usepackage{tabularx}

\begin{document}
\newcommand{\agent}{Almanac Copilot }

\title[Article Title]{\textit{\agent}: Towards Autonomous Electronic Health Record Navigation}


\author*[1]{\fnm{Zakka} \sur{Cyril}}\email{czakka@stanford.edu}
\author[1]{\fnm{Cho} \sur{Joseph}}
\author[2]{\fnm{Fahed} \sur{Gracia}}
\author[3]{\fnm{Shad} \sur{Rohan}}
\author[4]{\fnm{Moor} \sur{Michael}}
\author[1]{\fnm{Fong} \sur{Robyn}}
\author[1]{\fnm{Kaur} \sur{Dhamanpreet}}
\author[5]{\fnm{Ravi} \sur{Vishnu}}
\author[5]{\fnm{Aalami} \sur{Oliver}}
\author[6]{\fnm{Daneshjou} \sur{Roxana}}
\author[7]{\fnm{Chaudhari} \sur{Akshay}}

\author*[1]{\fnm{Hiesinger} \sur{William}}\email{willhies@stanford.edu}

\affil[1]{\orgdiv{Department of Cardiothoracic Surgery}, \orgname{Stanford Medicine}}
\affil[2]{\orgdiv{Department of Cardiovascular Medicine}, \orgname{Stanford Medicine}}
\affil[3]{\orgdiv{Division of Cardiovascular Surgery}, \orgname{Penn Medicine}}
\affil[4]{\orgdiv{Department of Computer Science}, \orgname{Stanford University}}
\affil[5]{\orgdiv{Byers Center for Biodesign}, \orgname{Stanford University}}
\affil[6]{\orgdiv{Department of Dermatology}, \orgname{Stanford Medicine}}
\affil[7]{\orgdiv{Radiology \& Integrative Biomedical Imaging Informatics}, \orgname{Stanford Medicine}}


\abstract{Clinicians spend large amounts of time on clinical documentation, and inefficiencies impact quality of care and increase clinician burnout. Despite the promise of electronic medical records (EMR), the transition from paper-based records has been negatively associated with clinician wellness, in part due to poor user experience, increased burden of documentation, and alert fatigue. In this study, we present Almanac Copilot, an autonomous agent capable of assisting clinicians with EMR-specific tasks such as information retrieval and order placement. On EHR-QA, a synthetic evaluation dataset of 300 common EHR queries based on real patient data, Almanac Copilot obtains a successful task completion rate of 74\% (n = 221 tasks) with a mean score of 2.45/3 (95\% CI:2.34-2.56). By automating routine tasks and streamlining the documentation process, our findings highlight the significant potential of autonomous agents to mitigate the cognitive load imposed on clinicians by current EMR systems.}

\maketitle

\section{Introduction}\label{sec1}
The introduction of electronic medical records (EMRs) represented a significant milestone in the evolution of healthcare technology. This transformation from paper-based records to digital formats has been largely completed, with more than 95\% of acute care facilities and 85\% of outpatient practices now utilizing basic or certified electronic health record (EHR) systems \cite{myrick2019national, adlermilstein2017ehr} in the United States. The impetus for this shift stemmed from growing concerns about the inconsistencies in healthcare quality across the United States, coupled with the belief in the potential of computer technology to address a looming crisis of medical errors. From 2001 to 2017 alone, office-based physician adoption of EHRs rose from 18\% to 85.9\%, driven in large part by federal incentives \cite{blumenthal2010meaningful} to promote the adoption and meaningful use of certified EHRs.

\begin{figure*}[t]
\centering
\includegraphics[width=\textwidth]{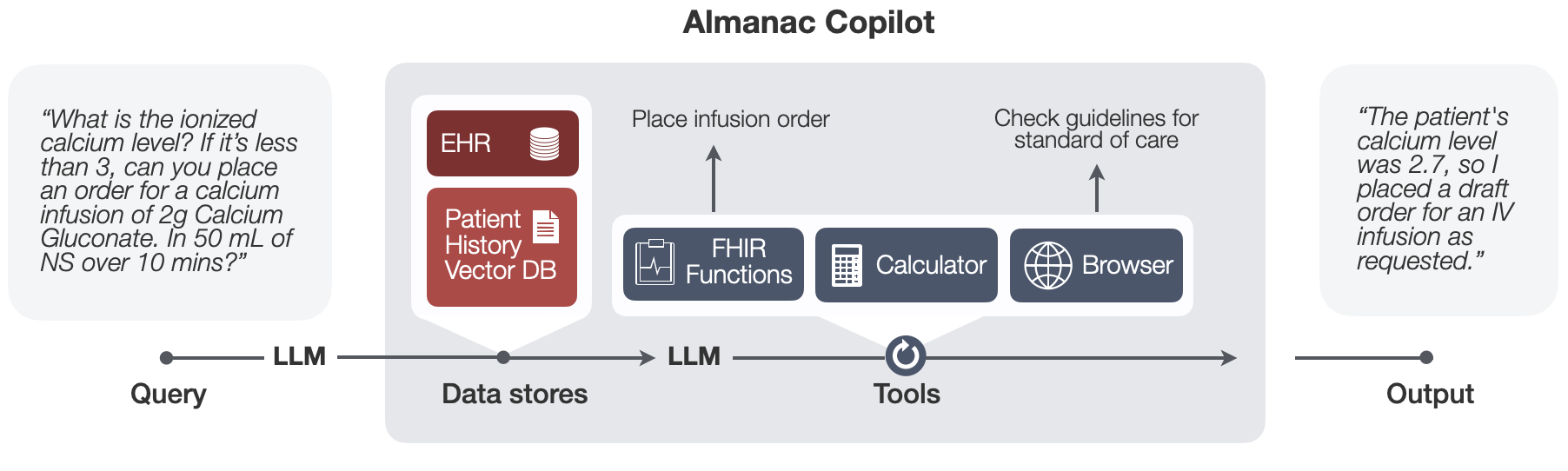} 
\captionsetup{width=\textwidth}
\caption{\textbf{Overview of the Almanac Copilot Architecture.} Upon receiving a query, the system dynamically selects a subset of APIs from a predetermined list of functions (i.e. FHIR functions, browser, calculator), optimizing the process to meet the specific requirements of the query.}
\label{overview}
\end{figure*}

However, despite their numerous advantages, the shift from paper to electronic records has inadvertently heightened stress levels among healthcare providers \cite{Kroth2018, Marmor2018, DelCarmen2019}, with nearly 75\% of clinicians with burnout symptoms pinpointing EHRs as a source \cite{Tajirian2020}. Factors contributing to this stress include intrinsic issues such as sub-optimal software design, and socio-technical factors, such as poor usability or workflow integration. EHRs have often compounded physicians’ cognitive load through excessive data entry requirements and diminished the time available for patient interaction \cite{Pelland2017}. This burnout has led to a cascade of negative outcomes, including increased rates of major medical errors, compromised quality of care, safety incidents, decreased patient satisfaction, and heightened turnover in the primary-care workforce \cite{Campbell2006, Shachak2009, Wachter2018, Friedberg2014}.

Recently, advances in generative artificial intelligence (AI), have raised the prospect of deploying effective and reliable tools to mitigate these challenges, with applications ranging from digital scribes \cite{vanBuchem2021DigitalScribe} to early clinical decision systems \cite{singhal2022large} and summarization tools \cite{VanVeen2023ClinicalText}. Although these developments show promise in controlled clinical settings \cite{Tierney2024AmbientAI}, they fall short in two key areas: first, their application is narrowly focused on isolated clinical encounters or highly curated retrospective datasets, which disregard the rich and comprehensive patient histories clinicians must take into account; second, they fail to address the multifaceted issues contributing to EHR fatigue, including extensive and redundant data entry, frequent interruptive alerts, poor integration with clinical and informatics workflows, as well as the absence of intuitive interfaces that resonate with the thought processes of clinicians.

In response to these challenges, we explore creating an autonomous EHR agent. Our objective is twofold: to reduce the cognitive burden on healthcare professionals and to improve the efficiency and effectiveness of healthcare delivery. We define three desiderata of an autonomous EHR:
\begin{itemize}
  \item \textbf{Information retrieval and summarization} focuses on streamlining access to patient data and medical knowledge, to reduce the time clinicians spend navigating complex EMR systems.
  \item \textbf{Data Manipulation} evaluates the system's ability to draft clinical notes, place orders for tests, medications, and other interventions. This process streamlines clinical workflows by reducing manual data entry.
  \item \textbf{Alert surfacing} assesses the system's effectiveness in prioritizing and presenting alerts to clinicians based on their relevance and urgency, while minimizing alert fatigue.
\end{itemize}

\noindent These core aspects delineate the spectrum of autonomy for the EHR agent, structured across three progressive levels:
\begin{itemize}
  \item \textbf{Level 0.} The clinician manually performs all tasks, with no assistance from the agent, representing the current standard practice.
  \item \textbf{Level 1.} The agent assists by preparing tasks based on explicit clinician commands, but the clinician reviews and approves all actions.
  \item \textbf{Level 2.} The agent proactively suggests actions based on contextual understanding and previous interactions, requiring clinician validation before placing them.
\end{itemize}

In this study, we introduce \textbf{Almanac Copilot}, a Level 1 autonomous agent framework aimed at enhancing the clinical workflow by alleviating the cognitive and administrative burdens on healthcare professionals. Rather than solely depending on large language models (LLMs) as lossy knowledgebases, we leverage their capabilities for \textit{knowledge extraction} and \textit{external tool utilization} embedded within the context of modern EHRs. In order to evaluate our framework, we curate \textbf{EHR-QA}, a synthetic dataset of 300 EHR-facing questions to closely mimic workflows often seen in care settings.

\begin{table*}[t]
\centering
\begin{tabularx}{\textwidth}{l|X|X}
\toprule
\textbf{Function Name} & \textbf{Description} & \textbf{Parameters} \\
\midrule
\multirow{4}{*}{\textbf{$search\_medical\_literature$}} 
                         & Return an answer (string) on diseases, treatment, symptoms, etc. from the medical literature 
                         & \begin{itemize}
                              \item query (str): The query to provide an answer to.
                            \end{itemize}\\
\hline
\multirow{11}{*}{\textbf{$create\_medication\_request\_order$}} 
                         & Creates a $MedicationRequest$ for the patient.
                         & \begin{itemize}
                              \item status (Enum): The current state of the medication request.
                              \item intent (Enum): Whether the request is a proposal, plan, or an original order.
                              \item name (string): Medication name.
                              \item ...
                            \end{itemize}\\
\hline
\multirow{13}{*}{\textbf{$search\_medication\_request\_database$}} 
                         & Fetches the appropriate $MedicationRequest$ history for each patient Encounter.  
                         & \begin{itemize}
                              \item query (string): The query to search for.
                              \item encounter\_id (string, optional): The ID of the $Encounter$ for which the $MedicationRequests$ were made.
                              \item date (string, optional): Date string to filter the returned $MedicationRequests$ with.
                            \end{itemize} \\
\bottomrule
\end{tabularx}
\caption{Sample functions provided to the Almanac Copilot to choose from in order to fulfill requests. The model is responsible for selecting, populating, and structuring the functions from the context provided.}
\label{tab:sample-funcs}
\end{table*}

While previous works have explored EHR-based agents in some capacity \cite{shi2024ehragent, vaid2024generative}, our approach takes a nuanced path that aligns more closely with the realities of modern healthcare informatics environments. We highlight our key contributions below:
\begin{itemize}
  \item \raggedright{\textbf{Fast Healthcare Interoperability Resources (FHIR) Compatibility \cite{HL7FHIROverview}}} 
  The FHIR standard is a set of rules and specifications for exchanging electronic health care data \cite{fhir}. Since its introduction in 2011, the standard has been widely adopted across all major EHR systems, and has even facilitated the ability for patients to visualize, understand and share their health data through mobile repositories \cite{schmiedmayer2024llm}.
  
  Moving away from the reliance on code generation for SQL database interactions - which may not only be misaligned with the current technological framework of hospitals but also pose risks to data security and integrity - we adopt the FHIR interoperability standard to ensure our framework is both robust and compatible with modern healthcare systems. 
 
  \item \textbf{Clinically-Aligned Benchmark} Although datasets already available in the literature \cite{lee2023ehrsql, wang2020text} may offer some value for broad medical inquiries or demographic-specific interventions, they fall short in benchmarking the most common processes of clinical workflows. To bridge this gap, we develop a clinician-derived synthetic benchmark comprising 300 questions across the core EHR tasks previously defined, ensuring relevance and applicability in real-world clinical settings. We detail our dataset generation workflow in section \ref{ehr-gen}.
  
  \item \textbf{Privacy First} In commitment to safeguarding patient data, our methodology is built on local model execution, optimized for consumer-grade GPUs. Unlike sending data to external servers via application processing interfaces (APIs), our approach guarantees that all sensitive information is processed within the secure environment of healthcare providers' systems. Local execution within institutional firewalls allows adhering to the highest standards of data privacy and security.
\end{itemize}

\section{Related Work}\label{relwork}
\noindent\textbf{LLMs and Tool Use.} The concept of tool learning in the realm of LLMs focuses on enhancing the abilities of these models by integrating them with various external functionalities via specific APIs \cite{qin2023toolllm, schick2023toolformer} or web browsers \cite{nakano2022webgpt}. This integration aims to extend the natural language processing capabilities of LLMs beyond mere text generation, enabling them to perform tasks that require interaction with external data sources or tools. Significant efforts have been made to equip LLMs with these capabilities, ranging from instruction fine-tuning \cite{yang2023gpt4tools} to exploration \cite{wang2024llms}.

\noindent\textbf{Clinical and Biomedical LLMs.} The development of LLMs for clinical and biomedical applications has seen a recent surge with the introduction of models like BioGPT \cite{Luo_2022}, SciBERT \cite{beltagy2019scibert}, NYUTron \cite{Jiang2023}, and MedPalm-2 \cite{singhal2023expertlevel}. These models have been used across a variety of clinical tasks, such as open-ended question-answering \cite{doi:10.1056/AIoa2300068}, clinical documentation \cite{VanVeen2023ClinicalText, yang2023customizing}, patient informed consent \cite{10.1001/jamanetworkopen.2023.36997}, International Classification of Disease (ICD) coding \cite{huang2022plmicd}, mental health support \cite{dechoudhury2023benefits}, and medical education \cite{Qiu_2023}.

\noindent\textbf{LLM Agents} There exists a rich body of works exploring the role of autonomous LLM agents that demonstrate capabilities closely mimicking those of autonomous design, planning, and execution \cite{wang2023voyager, hong2023metagpt, yao2023react, shen2023hugginggpt, yang2023autogpt, bran2023chemcrow, hu2023chatdb, qian2023communicative}. In the medical domain, these efforts have mostly been confined to a combination of population-based questions operating directly on Structured Query Language (SQL) databases \cite{shi2024ehragent} or open-ended medical question-answering on EHR extracted clinical scenarios \cite{vaid2024generative}.

\section{Methods}\label{methods}
\subsection{Architecture}
\agent is composed of a series of components working asynchronously to ensure accurate query completion (Figure \ref{overview}). An overview of each component is outlined below:
\subsubsection{Large Language Model}
The large language model architecture is a 33B parameter instruction-tuned Transformer decoder \cite{vaswani2023attention}, with major architectural improvements detailed below. We tailor our instruction-tuning dataset for tool use - or the ability for the model to select, populate, and chain pre-defined functions when presented with a query. These changes are tailored to optimize for computational efficiency on consumer-grade hardware without sacrificing downstream performance.

\begin{table*}[t]
\centering
\begin{tabularx}{\textwidth}{l|X|X}
\toprule
\textbf{Assessment Criteria} & \textbf{Question} & \textbf{Rationale} \\
\midrule
\multirow{3}{*}{\textbf{Functions Called (1 pt)}} 
                         & \textit{Are the functions called appropriate in nature to answer the given query?} 
                         & This axis examines the validity of the tools selected by the framework to address the user's query.\\
\hline
\multirow{5}{*}{\textbf{Parameter Choice (1 pt)}} 
                         & \textit{Are the parameters used in each of the functions derived from and technically valid for the given query?}
                         & Ensures the selection and application of parameters within each function are aligned with the requirements of the user's query and adhere to the function's definitions. \\
\hline
\multirow{3}{*}{\textbf{Script Validity (1 pt)}} 
                         & \textit{Can the provided answer be executed as a valid script?}
                         & Evaluates if the sequence of functions invoked by the framework forms a coherent and executable script.  \\
\bottomrule
\end{tabularx}
\caption{Criterion used to evaluate the \agent framework on the EHR-QA dataset of synthetic physician queries. These criteria are established to ensure that the model selects the correct tools, populates them with the correct parameters, and chains them in a valid order.}
\label{tab:guidelines}
\end{table*}

\noindent\textbf{Multi-Query Attention \cite{shazeer2019fast}.} At the core of the transformer architecture is Multi-Headed Attention (MHA), an attention mechanism that enables the model to focus on different parts of the input sequence simultaneously for various 'heads', improving its ability to capture complex relationships in the data. MHA is replaced with Multi-Query Attention (MQA) to reduce the memory requirement during decoding, allowing for higher batch sizes and higher inference throughput.

\noindent\textbf{Rotary Positional Embeddings (RoPE) \cite{su2023roformer}.} Position embeddings are traditionally added to the input sequences of transformers to convey relative and absolute token positioning to the model. The absolute positional embeddings is replaced with rotary positional embeddings, allowing for the flexibility to expand generation to any sequence length. 

\noindent\textbf{Normalizer Location \cite{zhang2019root}.} Both the input and output of each transformer sub-layer are normalized using RMSNorm to improve training stability and efficiency.

\noindent\subsubsection{Embedding Model}
Matryoshka Representation Learning (MRL) \cite{kusupati2024matryoshka} embeds information at multiple granularity levels, in a coarse-to-fine manner, within a single high-dimensional vector. To cater to various computing and latency regimens in the healthcare space, we make use of a flexible Matryoshka embedding model that can be adapted to multiple downstream tasks.

\subsubsection{Tools}
Several external tools are made available to \agent in order to automate complex tasks that require reasoning and decision-making across multiple systems. In order to ensure an optimal environment for task completion, functions are defined as self-contained modules, thus limiting the total number of functions available for selection to the model. Sample functions from a total of 9 pre-defined functions are provided in Table \ref{tab:sample-funcs} for reference. These tools are presented to the model as a schema describing the function's purpose, required parameters and associated descriptions, as well as the return type.

\noindent\textbf{Browser \cite{doi:10.1056/AIoa2300068}.} The browser is configured to retrieve and surface answers in response to clinical queries from a list of publicly available medical data repositories (e.g. PubMed).

\noindent\textbf{Calculators.} The clinical calculators are sourced from MDCalc, formatted in markdown, stored in the vector database, and evaluated inside a Python Read Evaluate Print Loop (REPL). 

\noindent\textbf{Database.} The database is a high-performance vector similarity engine optimized for the rapid indexing and search of embedded content through a combination of sparse and dense retrieval methods. We make use of Qdrant (v. 1.8.0) \cite{qdrant} for all experiments.

\noindent\textbf{Electronic Health Record (EHR) system.} A sandboxed version of an electronic health record system is populated with a combination of synthetic patient data \cite{Walonoski2018Synthea} and publicly available de-identified patient data  (e.g. MIMIC-IV \cite{goldberger2000physionet, johnson2020mimiciv}). Interactions with the database are made possible with carefully designed FHIR-based function calls to ensure compatibility with modern EHR infrastructures.

\begin{figure*}
     \centering
     \includegraphics[width=\textwidth, scale=0.1]{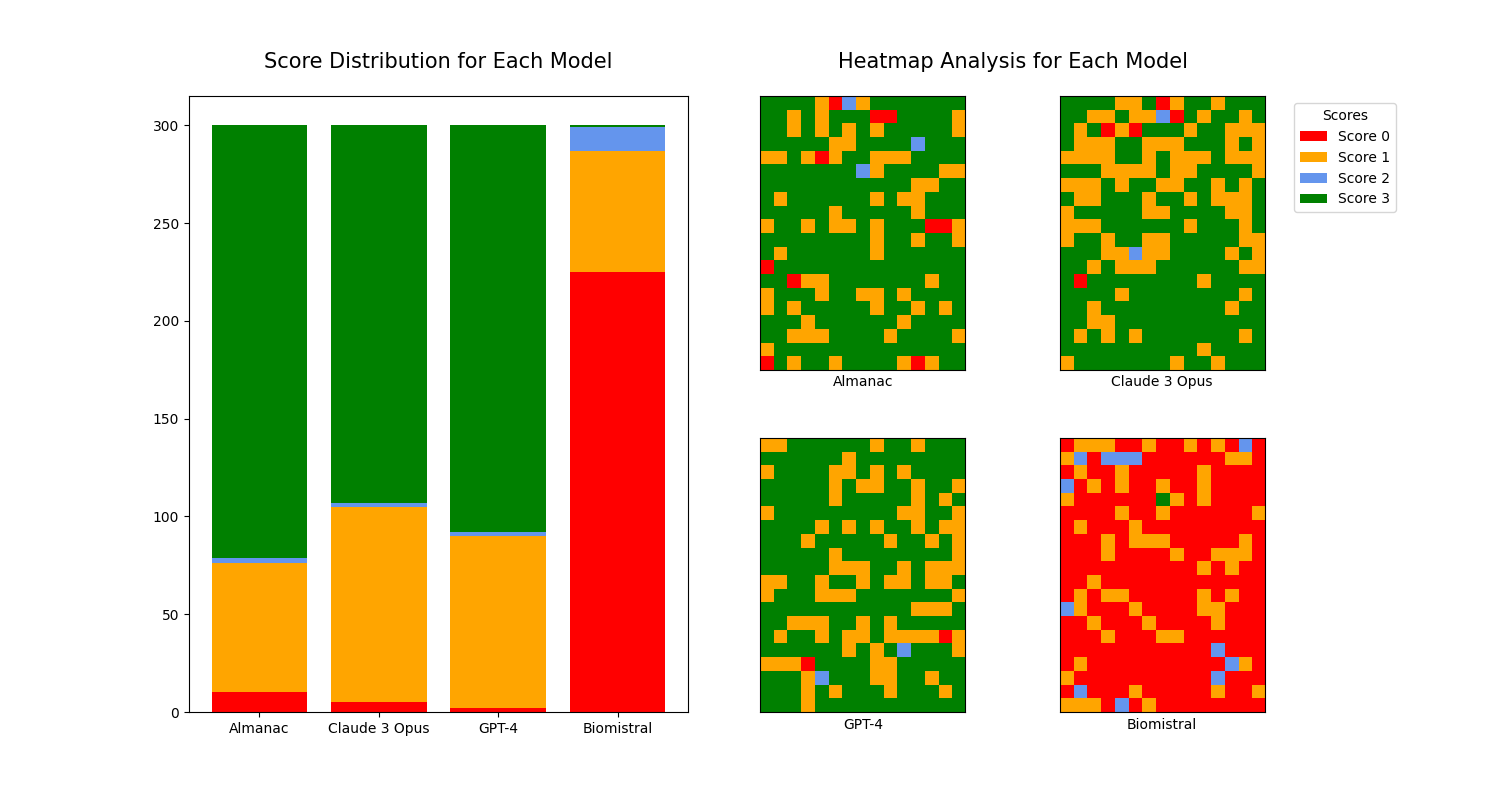}
     \caption{\textbf{Performance Evaluation of Almanac Copilot, ChatGPT-4, Claude 3 Opus and Biomistral on EHR-QA} a) The stacked bar plot illustrates the frequency of scores obtained across 300 synthetic questions within the EHR-QA framework. b) The heatmaps illustrate the models' performance in responding to the same dataset of questions. Each green square indicates a perfect score across all assessed metrics on the task in question. In contrast, red squares signal the lowest performance level. The scoring is sequential — subsequent correct actions are not credited if preceding steps are incorrect.}
     \label{fig:results}
 \end{figure*}

\subsection{EHR-QA Dataset Generation}\label{ehr-gen}
To robustly evaluate our framework across our previously defined task categories (i.e. information retrieval, summarization and data entry), we synthesize a novel dataset of 300 questions in a controlled and stepwise fashion using physician-generated template questions generated in a crowd-sourced fashion. \footnote{Following the protocols established by the MIMIC-IV \cite{goldberger2000physionet, johnson2020mimiciv} dataset, we provide access solely to the tools necessary for EHR-QA generation. For additional details, we direct readers to the associated code repository.}. To ensure a broad coverage of questions across our proxy dataset (MIMIC-IV), we create a series of template questions to capture common clinician use-cases within the context of EHR systems (Appendix \ref{secA1}). Constructing the final dataset is done iteratively for 300 rounds: at each round a patient is randomly selected from our pool of proxy EHR data, and an off-the-shelf LLM is used to generate a grounded question using a randomly selected template question (Positive Prompt). To simulate cases where clinicians might ask questions without clear answers, the model is also prompted to generate a question with no answer with a probability $p = 0.1$ (Negative Prompt). We use the following prompts:
\newline
\newline
\textbf{Positive Prompt:}
\textit{\say{Given the following template question (T): \{question\_template\}, and the following patient history (H): \{patient\_history\} generate a question based on T and H. Only use information present within H.}}
\newline
\newline
\textbf{Negative Prompt:}
\textit{\say{Given the following patient history (H): \{patient\_history\} generate a random medical question unrelated to (H).}}
\newline

\noindent The final dataset is manually inspected to ensure that the synthetic questions correlate with the information present in the full patient history.

\subsection{Evaluation}
The primary goal of this manuscript is to assess the efficacy of the \agent framework in carefully choosing, populating, and integrating the appropriate tools to respond to queries related to EHRs. Although the tools implemented are aptly set up to fulfill their designated functions as per their descriptions, it is important to note that other capabilities of LLMs across tasks such as summarization \cite{VanVeen2023ClinicalText} and retrieval-augmented generation \cite{doi:10.1056/AIoa2300068} have been empirically explored in other works and thus fall outside the scope of this analysis.

Instead, we objectively measure and report the nature, sequence, parameter choice, and overall execution validity of each of the functions called to answer the EHR-QA queries. In our assessment, we establish three benchmarks to deem a response successful, as outlined in Table \ref{tab:guidelines}, with each criterion met, earning 1 point. Scoring is sequential — subsequent correct actions are not credited if preceding steps are incorrect. We run all local experiments and models on a machine comprised of 4 NVIDIA Quadro RTX A5000 24GB GPUs - servers that can feasibly be deployed in clinical settings. As baselines, we also report the performances of ChatGPT-4 (Version: \textit{gpt-4-turbo-preview}, Accessed April 12, 2024) \cite{openai2024gpt4}, Claude 3 Opus (Version: \textit{claude-3-opus-20240229}, Accessed April 12, 2024)  \cite{Anthropic2023Claude3}, and BioMistral \cite{labrak2024biomistral} on EHR-QA, along with their mean scores. 

\section{Results}
In this section, we provide a summary of our results (Figure \ref{fig:results}).

Across the EHR-QA dataset, \agent obtains a success rate of 74\% (3 points; n = 221 tasks), with partial successes of 1\% (2 points; n = 3) and 22\%  (1 point; n = 66) respectively. We observe complete failure (0 points) in 3\% of cases, across 10 tasks, with an overall mean performance of $2.45$ (95\% CI: $2.34$--$2.56$).

Similar performances are observed for Opus and ChatGPT-4, with mean performances of $2.28$ (95\% CI: $2.17$--$2.39$) and $2.39$ (95\% CI: $2.28$--$2.49$) respectively. Opus obtains a complete success rate of 64\% (3 points; n = 193), followed by partial successes of 1\% (2 points; n = 2) and 33\% (1 point; n = 100), with a complete failure rate of 2\% (0 points; n = 5). GPT-4 performed comparably, with 69\% (3 points; n = 208), 1\% (2 points; n = 2), 29\% (1 point; n = 88), and 1\% (0 points; n = 2). On the other hand, BioMistral performed poorly with a mean score of $0.30$ (95\% CI: $0.23$--$0.36$), with a score distribution of 0\% (3 points; n = 1), 4\% (2 points; n = 12), 21\% (1 point; n = 62), and 75\% (0 points; n = 225).

We note that the majority of failures for the top three performing models occur as a result of hallucinations - or the erroneous generation of information without any basis in the provided query or context data within the parameters. These hallucinations often manifested as incorrect medication indications, fabricated tools, or the confounding of patient ID numbers with International Organization for Standardization (ISO) format dates.


\section{Discussion}
Today, clinicians are tasked with navigating, filtering, and effectively utilizing an extensive array of tools, with actions often centralized around EHR systems. Despite the advantages offered by these systems in streamlining administrative tasks, they have paradoxically contributed to the increasing burnout levels among healthcare professionals in the United States.

To address this issue, numerous studies have explored the application of generative AI in medicine, aiming to enhance knowledge transfer, documentation, and clinical summarization through open-ended question-answering \cite{doi:10.1056/AIoa2300068}, clinical text summarization \cite{VanVeen2023ClinicalText}, and digital scribing \cite{vanBuchem2021DigitalScribe}. While these solutions offer benefits in specific areas, they tend to overlook the broader, intricate challenges at the heart of clinical workflows.

In this paper, we present \emph{\agent}, a Level 1 autonomous EHR agent designed to mitigate the cognitive and administrative burdens imposed on healthcare professionals by contemporary EMR systems. Through extensive evaluations on EHR-QA, a proxy dataset of common EHR tasks, Almanac Copilot achieves a success rate of 74\% matching the performance of models with parameter counts orders of magnitude larger than it, highlighting its potential to streamline clinical workflows and reduce the time clinicians dedicate to navigating complex interfaces. Although previous works such as Shi et. al. \cite{shi2024ehragent} have explored LLM-based EHR QA before, our approach differs 1) in its ability to natively and securely interface with the majority of modern EHR and healthcare infrastructures, 2) and its clinically-focused evaluation benchmark. 

Despite the promising outcomes showcased by our framework, critical hurdles must be addressed before clinical deployment. Foremost among these challenges is the potential for the agent to hallucinate, particularly when faced with incomplete data inputs from users. While strategies such as prompt engineering, supervised fine-tuning (SFT), or reinforcement learning with human feedback (RLHF) have shown potential in reducing these behaviors, they do not entirely eliminate them, and may paradoxically negatively impact model behavior. Additionally, despite serving as a robust benchmark for common EMR tasks, EHR-QA fails to capture the diversity and distribution of real-world clinical queries, which may seek information beyond the scope of a well-curated dataset such as Med-QA. Finally, the dataset's focus on single-hop or dual-hop reasoning tasks also falls short of encompassing the complexity of clinician queries, which often require more intricate reasoning and synthesis of information from multiple sources.

To move forward, several key enhancements are envisioned for the development of Level 2 autonomous EHR agents and their deployment in safety-critical settings. These include refining the agents' ability to understand and retain context over multiple interactions \cite{moniz2024realm}, reducing inference times and latency, and expanding capabilities to interpret image-based medical modalities (e.g. computed tomography scans, magnetic resonance imaging, and X-rays) \cite{openai2024gpt4, moor2023medflamingo} via specialized model calls. We believe that these advancements are essential for reducing the cognitive burden of clinicians, while bridging the gap between the current capabilities of \agent and the demands of real-world practice, paving the way for friction-less deployment in healthcare settings.

\section{Acknowledgments}
We would like to thank Graham Walker from \href{https://www.mdcalc.com}{MDCalc} for their permission to include clinical calculators in our vector database for the duration of our experiments. We would also like to thank Robert Brooks and the \href{https://lambdalabs.com}{Lambda Labs} team for supporting our experiments with GPU access.

\section*{Declarations}
\subsection{Funding}
This project was supported in part by a National Heart, Lung, and Blood Institute (NIH NHLBI) grant (1R01HL157235-01A1) (W.H.).

\subsection{Competing interests}
The authors declare no competing interests.

\subsection{Authors' contributions}
C.Z., A.C., and W.H. designed the experiments. C.Z. wrote the manuscript. Experimentation and manuscript feedback was given by R.D., O.A, V.R, A.C., M.M. and W.H. The code-base was authored by C.Z. and J.C. Computational experiments were performed by C.Z. and J.C. Evaluations were carried out by C.Z., J.C., and G.F. Template questions for EHR-QA were curated by G.F. and R.S. The work was supervised by W.H.







\onecolumn
\pagebreak
\begin{appendices}
\section{Template Questions for Synthetic Data Generation}\label{secA1}

\begin{table}[h!]
\centering
\begin{tabularx}{\textwidth}{c|X}
\toprule
\textbf{Type} & \textbf{Question} \\
\midrule
\multirow{29}{*}{Information Retrieval} 
                         & What were the results of $<test>$ on $<date>$? \\
                         & Were any of $<date>$'s results abnormal? \\
                         & Is $<test>$ value abnormal? \\
                         & What is the normal range for $<test>$? \\
                         & Did $<patient>$ ever have $<test>$ done? What was the value? \\
                         & Did $<patient>$ ever have $<imaging>$ done? What was the result? \\
                         & What is $<patient>$'s historical range for $<test>$? \\
                         & Can you summarize $<patient>$'s test results for $<date>$? \\
                         & What tests are still pending? \\
                         & Can you summarize $<patient>$'s medical history? \\
                         & Can you summarize $<patient>$'s last encounter? \\
                         & How was $<patient>$'s $<condition>$ previously controlled? \\
                         & Has $<patient>$ ever been diagnosed or treated for $<condition>$? \\
                         & Has $<patient>$ ever taken $<medication>$? On what date? And what for? \\
                         & How long has $<patient>$ been on $<medication>$ for? \\
                         & Does $<patient>$ have any recorded side-effects for $<medication>$? \\
                         & What dosage of $<medication>$ is $<patient>$ taking? \\
                         & What medications is patient currently taking? List dosage and frequency \\
                         & How often does $<patient>$ take $<medication>$? \\
                         & When was the last time $<patient>$ took $<medication>$? \\
                         & What is $<patient>$'s family and social history? \\
                         & Does $<patient>$ have any allergies? \\
                         & When did $<patient>$ have $<procedure>$? \\
                         & Did $<specialty>$ ever see $<patient>$? \\
                         & Did $<specialty>$ ever see $<patient>$ for $<condition>$? \\
                         & Can you summarize $<patient>$'s discharge note? \\
                         & What medications was $<patient>$'s discharged with? \\
                         & Can you lookup the treatment guidelines for $<condition>$? \\
                         & Has $<patient>$ been screened for $<condition>$? \\

\hline
\multirow{11}{*}{Data Entry} 
                         & Can you place an order for $<test>$? \\
                         & Can you place an order for $<imaging>$ with indication $<message>$ ? \\
                         & Can you place an order for $<medication>$ with $<dosage>$ for $<frequency>$? \\
                         & Can you place an order for a consult from $<specialty>$ for $<condition>$? \\
                         & Can you place an order for vitals for $<patient>$  for $<frequency>$? \\
                         & If $<patient>$'s $<test>$ is $<below/above>$ $<value>$, can you place an order for $<medication>$ with $<dosage>$ for $<frequency>$?  \\
                         & Can you place a discharge order for $<patient>$? \\
                         & Can you notify nurse that $<message>$? \\
                         & Can you place a transfer order to $<department>$? \\
                         & Can you place an NPO order for $<patient>$ after midnight?  \\
                         & Can you place an order to consent $<patient>$ for $<procedure>$?  \\
\bottomrule
\end{tabularx}
\caption{In order to create the EHR-QA evaluation benchmark, a series of 40 template questions based on common clinician use cases are manually curated. An off-the-shelf LLM is used for \textit{grounded question generation}, using the patient's history as context for a total of 300 questions.}
\label{tab:questions}
\end{table}

\end{appendices}


\bibliography{sn-bibliography}

\end{document}